\documentclass{article}

 \usepackage[preprint]{neurips_2025}

\usepackage[utf8]{inputenc} % allow utf-8 input
\usepackage[T1]{fontenc}    % use 8-bit T1 fonts
\usepackage{hyperref}       % hyperlinks
\usepackage{url}            % simple URL typesetting
\usepackage{booktabs}       % professional-quality tables
\usepackage{amsfonts}       % blackboard math symbols
\usepackage{nicefrac}       % compact symbols for 1/2, etc.
\usepackage{microtype}      % microtypography
\usepackage{xcolor}         % colors
\usepackage{graphicx}       % Required for inserting images
\usepackage{amsmath}        % For mathematical equations
\usepackage{listings}       % For code listings
\usepackage{multirow}       % For multi-row cells in tables
\usepackage{pgfplots}       % For plotting
\usepackage{tikz}           % For plotting
\usepackage{siunitx}        % For aligning numbers by decimal point
\usepackage{caption}        % For customizing captions

\pgfplotsset{compat=1.18}    % Sets pgfplots compatibility

\definecolor{codegreen}{rgb}{0,0.6,0}
\definecolor{codegray}{rgb}{0.5,0.5,0.5}
\definecolor{codepurple}{rgb}{0.58,0,0.82}
\definecolor{backcolour}{rgb}{0.95,0.95,0.92}

\lstdefinestyle{mystyle}{
    backgroundcolor=\color{backcolour},   
    commentstyle=\color{codegreen},
    keywordstyle=\color{magenta},
    numberstyle=\tiny\color{codegray},
    stringstyle=\color{codepurple},
    basicstyle=\footnotesize\ttfamily,
    breakatwhitespace=false,         
    breaklines=true,                 
    captionpos=b,                    
    keepspaces=true,                 
    showspaces=false,                
    showstringspaces=false,
    showtabs=false,                  
    tabsize=2
}
\title{HEFT: A Coarse-to-Fine Hierarchy for Enhancing the Efficiency and Accuracy of Language Model Reasoning}

\author{
  Brennen Hill \\
  Department of Computer Science\\
  University of Wisconsin-Madison\\
  Madison, WI 53706 \\
  \texttt{bahill4@wisc.edu} \\
}

\begin{document}

\maketitle

\begin{abstract}
The adaptation of large language models (LLMs) to specialized reasoning tasks is fundamentally constrained by computational resources. Parameter-Efficient Fine-Tuning (PEFT) methods have emerged as a powerful solution, yet the landscape of these techniques is diverse, with distinct methods operating in either the model's weight space or its representation space. This paper investigates the hypothesis that a synergistic combination of these paradigms can unlock superior performance and efficiency. We introduce HEFT (Hierarchical Efficient Fine-Tuning), a novel hierarchical adaptation strategy that composes two distinct PEFT methods in a coarse-to-fine manner: first, a broad, foundational adaptation in the weight space using Low-Rank Adaptation (LoRA), followed by a precise, surgical refinement of internal activations using Representation Fine-Tuning (ReFT). We evaluate this approach by fine-tuning a Llama-2-7B model on the BoolQ benchmark, a challenging dataset for inferential reasoning. Our results reveal a profound synergistic effect. A model fine-tuned for only three epochs with our HEFT strategy achieves an accuracy of 85.17\%, exceeding the performance of models trained for 20 epochs with either LoRA-only (85.05\%) or ReFT-only (83.36\%) methodologies. This work demonstrates that the thoughtful composition of PEFT methods is a potent algorithmic innovation, offering a more efficient and effective path toward advancing the reasoning capabilities of language models. By achieving superior results with a fraction of the computational budget, our findings present a principled approach to overcoming the obstacles inherent in adapting large-scale models for complex cognitive tasks.
\end{abstract}

\section{Introduction}

The advent of large language models (LLMs) has revolutionized natural language processing, largely driven by a transfer learning paradigm where massive models are pre-trained on web-scale data and subsequently adapted to downstream tasks \citep{liu2022few}. While effective, the standard adaptation method of full-parameter fine-tuning is exceptionally resource-intensive, creating a significant barrier to both research and practical deployment \citep{hu2021lora}. This has catalyzed a vibrant area of research into Parameter-Efficient Fine-Tuning (PEFT), a family of techniques that adapt pre-trained models by updating only a small, strategically chosen subset of parameters \citep{zhao2025peftsurvey}.

Within the PEFT landscape, two prominent and conceptually distinct paradigms have gained traction. The first, Low-Rank Adaptation (LoRA), operates in the model's \textit{weight space}. Based on the hypothesis that weight updates during adaptation have a low intrinsic rank, LoRA freezes the original weights and injects small, trainable low-rank matrices to approximate the task-specific update \citep{hu2021lora}. The second paradigm, Representation Fine-Tuning (ReFT), operates in the model's \textit{representation space}. Drawing insights from interpretability research, ReFT keeps the entire base model frozen and instead learns to apply direct, surgical interventions on the model's hidden representations as they flow through the network \citep{wu2024reft}.

While both LoRA and ReFT offer dramatic efficiency gains, their mechanisms suggest complementary strengths. LoRA enacts a broad, foundational shift in the model's parameterization but has been shown to introduce structural artifacts that can be linked to forgetting \citep{sharma2024lora}. ReFT offers highly precise and interpretable edits to semantic pathways, proving exceptionally effective for commonsense reasoning tasks, but may be less suited for steering long-form generative processes than global weight-space methods \citep{wu2024reft}. This distinction raises a crucial question: can these disparate methodologies be combined to yield a result greater than the sum of their parts?

This paper introduces and validates HEFT (Hierarchical Efficient Fine-Tuning), a novel hierarchical adaptation strategy based on a coarse-to-fine hypothesis. We propose first applying LoRA for broad, foundational tuning, followed by ReFT for targeted, high-precision refinement. We test this approach on the BoolQ benchmark, a question-answering dataset specifically designed to probe for complex, entailment-like inference \citep{clark2019boolq}. Our empirical results are striking:
\begin{itemize}
  \item Fine-tuning a Llama-2-7B model with just \textbf{three epochs} of our HEFT strategy achieves an accuracy of \textbf{85.17\%} on the BoolQ validation set.
  \item This result surpasses the accuracy achieved after a full \textbf{20 epochs} of training with LoRA alone (85.05\%) or ReFT alone (83.36\%).
  \item A 20-epoch HEFT approach reaches an even higher accuracy of \textbf{85.47\%}, demonstrating that the synergistic benefits are sustained with further training.
\end{itemize}

These findings provide compelling evidence that combining weight-space and representation-space adaptation is a highly effective and efficient strategy. Our hierarchical fine-tuning method, HEFT, is a principled \textbf{algorithmic innovation} that overcomes a \textbf{fundamental challenge}, the trade-off between performance and computational cost in model adaptation. By demonstrating how a smaller model can achieve reasoning capabilities competitive with models an order of magnitude larger, we illuminate a pathway to more efficiently improving how language models \textbf{solve complex tasks}.

\section{Related work}

\subsection{The imperative for parameter-efficient fine-tuning}
The standard practice of full-parameter fine-tuning (FT), which updates every weight in a pre-trained model, poses prohibitive scalability challenges. The computational and memory costs are immense, requiring high-end GPU clusters and extensive training time, placing it beyond the reach of many researchers \citep{hu2021lora}. The memory footprint for optimizer states, gradients, and activations can be several times the size of the model itself \citep{siddika2025fedreft}. Furthermore, FT introduces significant logistical burdens, as it produces a complete, multi-gigabyte copy of the model for each downstream task. A more fundamental drawback is catastrophic forgetting, where the intense adaptation to a narrow task can cause the model to lose some of the general linguistic knowledge acquired during pre-training \citep{sharma2024lora}.

In response, PEFT methods have emerged as a transformative solution \citep{zhao2025peftsurvey}. By freezing the vast majority of a model's weights and training only a small subset of new or existing parameters, PEFT dramatically reduces computational demands, lowers memory requirements, and accelerates training cycles \citep{liu2022few}. The landscape of PEFT is diverse, including additive methods like Adapters \citep{houlsby2019parameter}, which inject small new modules; selective methods like BitFit \citep{zaken2021bitfit}, which tune only bias terms; and reparameterization-based methods, of which LoRA is the most prominent example \citep{fu2023effectiveness}.

\subsection{Weight-space adaptation: Low-rank adaptation (LoRA)}
LoRA, introduced by \citet{hu2021lora}, has become one of the most widely adopted PEFT techniques. Its foundation is the low intrinsic rank hypothesis, which posits that the change in a model's weight matrix, $\Delta W$, during adaptation can be effectively approximated by a matrix of much lower rank. LoRA operationalizes this by freezing the pre-trained weights $W_0$ and representing the update as a low-rank product, $\Delta W = BA$. Here, $A \in \mathbb{R}^{r \times k}$ and $B \in \mathbb{R}^{d \times r}$, with the rank $r \ll \min(d, k)$.

During training, only the parameters of $A$ and $B$ are updated. The modified forward pass is computed as:
$$h = W_0x + \Delta Wx = W_0x + BAx$$
To control the magnitude of the adaptation, the update is often scaled by a hyperparameter $\alpha$, resulting in $h = W_0x + \frac{\alpha}{r}BAx$. This reparameterization can reduce the number of trainable parameters by a factor of 10,000 \citep{hu2021lora}. A key practical advantage of LoRA is that it introduces no inference latency, as the learned matrices $A$ and $B$ can be multiplied and merged with the original weight matrix $W_0$ after training is complete.

The foundational success of LoRA has spurred an ecosystem of advanced variants. LoRA+ improves performance and training speed by using different learning rates for the $A$ and $B$ matrices \citep{soufiane2024lora+}. Recognizing that a fixed rank for all layers is suboptimal, AdaLoRA \citep{zhang2023adalora} and SalientLoRA \citep{zhang2024salientlora} introduce methods for dynamically allocating the rank budget to layers based on importance scores, leading to more efficient and effective parameter allocation.

Despite its success, LoRA is not without limitations. A critical analysis by \citet{sharma2024lora} revealed that LoRA's parameter updates are structurally distinct from those of full fine-tuning. They find that LoRA introduces intruder dimensions, new, high-ranking singular vectors in the adapted weight matrix that are dissimilar to any vectors in the pre-trained model. These dimensions are causally linked to the forgetting of the pre-training distribution, providing a mechanistic explanation for some of LoRA's shortcomings and motivating the exploration of complementary fine-tuning paradigms.

\subsection{Representation-space intervention: Representation fine-tuning (ReFT)}
A conceptually distinct paradigm, Representation Fine-Tuning (ReFT), shifts the focus from modifying model weights to directly editing the model's internal hidden representations \citep{wu2024reft}. This approach is born from interpretability research, which has shown that LLM activations encode rich, structured semantic information. The central hypothesis of ReFT is that directly manipulating these meaningful representations is a more powerful and efficient mechanism for steering model behavior than the indirect method of updating weights.

ReFT operates on a completely frozen base model, learning task-specific intervention functions that are applied to a subset of hidden states at specific layers and token positions during the forward pass. The most prominent instantiation, Low-rank Linear Subspace ReFT (LoReFT), is built on the insight that high-level concepts are often encoded within linear subspaces of the representation space. LoReFT learns a low-rank projection matrix $R$ and a linear transformation $(W, b)$ that operates within that subspace. The formal intervention on a hidden representation $h$ is:
$$ \Phi_{\text{LoReFT}}(h) = h + R^T((W(Rh)+b) - Rh) $$
This mechanism is exceptionally parameter-efficient, learning interventions that are 15 to 65 times smaller than a typical LoRA configuration while often achieving superior performance, particularly on commonsense reasoning benchmarks \citep{wu2024reft}. As a newer paradigm, ReFT presents active research frontiers, including its adaptation to challenging settings like Federated Learning, which required the development of novel aggregation strategies like in FedReFT \citep{siddika2025fedreft}, and addressing multi-skill interference through compositional methods like CS-ReFT \citep{pan2025csreft}.

\subsection{The BoolQ benchmark for inferential reasoning}
Meaningful evaluation of reasoning capabilities requires benchmarks that test more than factual recall. The Boolean Questions (BoolQ) dataset was created to provide such a challenge \citep{clark2019boolq}. It consists of 15,942 naturally occurring yes/no questions sourced from unprompted user search queries, paired with a relevant passage from a Wikipedia article.

The dataset's surprising difficulty stems from its naturalistic origin. Unlike extractive QA datasets, BoolQ questions often require models to perform complex, entailment-like inference, synthesizing information from multiple sentences and leveraging world knowledge to connect the passage's content to the question's premise. The original study found that transfer learning from a large-scale natural language inference (NLI) dataset was the most effective strategy for improving BoolQ performance, providing strong evidence that the core competency being tested is indeed inferential reasoning \citep{clark2019boolq}. Its inclusion as a core task in the more rigorous SuperGLUE benchmark further solidifies its status as a robust measure of a model's deeper language understanding and reasoning abilities \citep{wang2019superglue, huang2023towards}.

\subsection{Composition of PEFT methods}
The proliferation of diverse PEFT methods has naturally led to research into their composition. Most existing work has focused on combining homogeneous modules, typically multiple LoRA adapters, under the umbrella of LoRA fusion or model merging \citep{asadi2024doescombiningparameterefficientmodules}. Simple approaches involve averaging the weights of adapters trained on different tasks. More sophisticated techniques learn optimal weightings for a linear combination of adapters or perform context-dependent fusion based on the input prompt. While promising, these methods combine modules of the same type. Our work explores a novel form of heterogeneous composition, hierarchically layering distinct PEFT paradigms to create a coarse-to-fine adaptation curriculum.

\section{Methodology}
Our approach, HEFT (Hierarchical Efficient Fine-Tuning), is a hierarchical, two-stage fine-tuning process designed to leverage the complementary strengths of weight-space and representation-space adaptation. The core hypothesis is that an initial coarse-grained adaptation with LoRA provides a strong foundational parameterization that can be more effectively and efficiently refined by a subsequent fine-grained intervention with ReFT.

\subsection{The coarse-to-fine rationale}
We frame the hierarchical application of LoRA followed by ReFT within the HEFT framework as a coarse-to-fine adaptation strategy.
\begin{enumerate}
  \item \textbf{Stage 1: Coarse-Grained Adaptation (LoRA).} By operating directly on the model's weight space, LoRA enacts a broad, foundational shift in the model's parameterization. This stage effectively moves the entire model into a region of the high-dimensional solution space that is better aligned with the general characteristics of the target task. For a reasoning task like BoolQ, this involves adapting the weights to better handle the inferential and logical style of the questions. This gets the model into the correct general semantic neighborhood but may lack precision.
  \item \textbf{Stage 2: Fine-Grained Refinement (ReFT).} After LoRA has completed its broad adaptation, ReFT performs a more targeted and surgical intervention. Operating on the model's internal representations, ReFT can directly edit the semantic pathways involved in the reasoning process. This allows for high-precision steering of the model's activations to refine its behavior and correct for any imprecision from the initial LoRA tuning. ReFT's demonstrated strength on commonsense reasoning makes it an ideal tool for this refinement stage \citep{wu2024reft}.
\end{enumerate}
This two-stage process offers a compelling path to achieving a better trade-off between efficiency and performance, a central theme in the pursuit of efficient reasoning models \citep{sui2025stop}.

\subsection{Experimental setup}
\paragraph{Base model} We use the \texttt{meta-llama/Llama-2-7b-chat-hf} model \citep{touvron2023llama}, a widely adopted 7-billion parameter LLM, as the foundation for all experiments.

\paragraph{Dataset} All fine-tuning and evaluation are performed on the BoolQ dataset \citep{clark2019boolq}. We use the official training split containing 9,427 examples for fine-tuning and report accuracy on the 3,270 examples in the validation split. Each example is formatted into a prompt instructing the model to answer a given question based on a provided passage with only "Yes" or "No".

\paragraph{Hardware and software} Experiments were conducted using the University of Wisconsin-Madison's Center for High Throughput Computing (CHTC). We utilized nodes running CentOS 8 Stream with the SLURM scheduler, each equipped with a single NVIDIA GPU with at least 32GB of VRAM.

\subsection{Implementation details}
Our proposed fine-tuning procedure consists of the following two stages:

\paragraph{Stage 1: LoRA fine-tuning}
We adapt the base Llama-2-7B model using LoRA. The LoRA configuration targets the model's linear layers with a rank ($r$) of 8 and a scaling factor ($\alpha$) of 32. The model is trained on the BoolQ training set for a specified number of epochs. Upon completion, the LoRA adapter weights are merged into the base model's weights, producing a new, consolidated model with no inference latency overhead.

\paragraph{Stage 2: ReFT fine-tuning}
We take the LoRA-adapted model from Stage 1 as the new base model. Its weights are then kept entirely frozen. We introduce a trainable Low-rank Linear Subspace ReFT (LoReFT) intervention. Specifically, we configure a 4-dimensional intervention on the block output representation of the 15th transformer layer. This lightweight intervention is then trained on the same BoolQ training set.

\subsection{Baselines and evaluation}
To validate the effectiveness of our hierarchical HEFT approach, we compare it against several baselines:
\begin{itemize}
  \item \textbf{LoRA-Only (20 epochs)}: The base model is fine-tuned for 20 epochs using only the LoRA configuration from Stage 1.
  \item \textbf{ReFT-Only (20 epochs)}: The base model is fine-tuned for 20 epochs using only the ReFT configuration from Stage 2.
  \item \textbf{HEFT (20+20 epochs)}: Our full hierarchical method, where both stages are trained for 20 epochs each, to measure the maximum performance of the combined strategy.
  \item \textbf{HEFT (3+3 epochs)}: Our primary experimental condition to test the efficiency of the hierarchical approach, where each stage is trained for only 3 epochs.
\end{itemize}
The primary metric for evaluation is accuracy on the BoolQ validation set.

\section{Results}
Our experiments confirm the hypothesis that HEFT, our hierarchical LoRA-then-ReFT fine-tuning strategy, yields significant benefits in both performance and efficiency. The results, summarized in Table \ref{tab:main_results}, demonstrate a clear synergistic effect where the combined approach outperforms its individual components, often with significantly less training.

\begin{table}[h!]
\centering
\caption{Performance on the BoolQ validation set. The hierarchical HEFT approach achieves superior accuracy with substantially fewer training epochs, highlighting a strong synergistic effect.}
\label{tab:main_results}
\begin{tabular}{@{}lcccc@{}}
\toprule
\textbf{Method} & \textbf{LoRA Epochs} & \textbf{ReFT Epochs} & \textbf{Accuracy (\%)} & \textbf{Time (H:M)} \\ \midrule
ReFT-Only & 0 & 20 & 83.36 & 2:19 \\
LoRA-Only & 20 & 0 & 85.05 & 6:52 \\ \midrule
\textbf{HEFT} & \textbf{3} & \textbf{3} & \textbf{85.17} & \textbf{1:23} \\
\textbf{HEFT} & \textbf{20} & \textbf{20} & \textbf{85.47} & \textbf{9:11} \\ \bottomrule
\end{tabular}
\end{table}

\subsection{Analysis of efficiency and performance synergy}
The most striking result is the performance of our efficient HEFT configuration. With only 3 epochs for the LoRA stage and 3 epochs for the ReFT stage, this method achieves an accuracy of \textbf{85.17\%}. This performance surpasses that of both the LoRA-only model trained for 20 epochs (85.05\%) and the ReFT-only model trained for 20 epochs (83.36\%). This key finding demonstrates that the combined approach not only reaches a better performance ceiling but does so with a fraction of the computational effort.

\begin{figure}[h!]
\centering
\begin{tikzpicture}
\begin{axis}[
  title={Accuracy vs. compute trade-off on BoolQ},
  xlabel={Training Time (minutes)},
  ylabel={Accuracy (\%)},
  xmin=0, xmax=600,
  ymin=82.5, ymax=87,
  grid=major,
  legend pos=south east,
  scatter/classes={
    a={mark=*, blue},
    b={mark=square*, red},
    c={mark=triangle*, brown},
    d={mark=diamond*, teal}
  }
]
\addplot[scatter, only marks, scatter src=explicit symbolic]
table [meta=label] {plot_data.dat};
\node[pin=85:{\shortstack[r]{\textbf{HEFT}\\\textbf{3+3 epochs}}}] at (axis cs:83, 85.17) {};
\node[pin=90:{\shortstack[l]{ReFT-Only\\20 epochs}}] at (axis cs:139, 83.36) {};
\node[pin=270:{\shortstack[l]{LoRA-Only\\20 epochs}}] at (axis cs:412, 85.05) {};
\node[pin=135:{\shortstack[l]{HEFT\\20+20 epochs}}] at (axis cs:551, 85.47) {};
\end{axis}
\end{tikzpicture}
\caption{The performance-efficiency frontier. Our efficient HEFT method (blue) achieves high accuracy for a fraction of the training time, occupying the desirable top-left region of the plot.}
\label{fig:accuracy_vs_compute}
\end{figure}
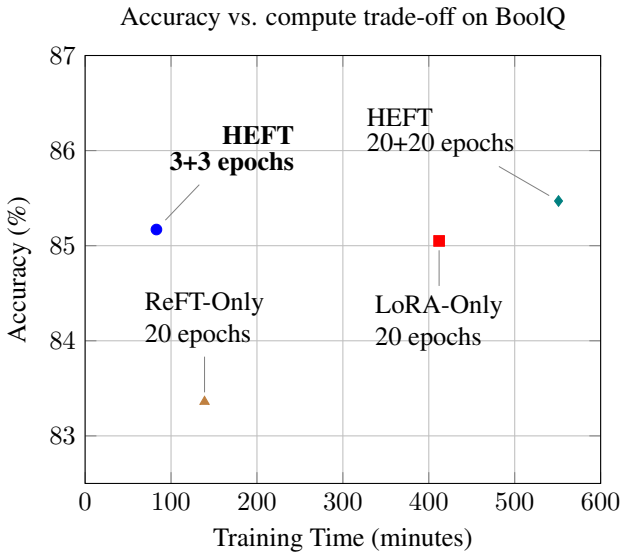

The training time further underscores this efficiency gain, as visualized in Figure \ref{fig:accuracy_vs_compute}. Our efficient 3+3 epoch HEFT run completed in 1 hour and 23 minutes, achieving a superior result in significantly less time than the LoRA-only (6 hours and 52 minutes) or ReFT-only (2 hours and 19 minutes minutes) baselines. This provides strong evidence that this is not merely an additive improvement but a multiplicative one, where the coarse-grained adaptation from the first stage makes the fine-grained refinement of the second stage dramatically more effective.

\subsection{Performance in the broader context}
To contextualize our results, we compare the performance of our fine-tuned 7B parameter model against other models on the BoolQ benchmark. As shown in Table \ref{tab:leaderboard_comparison}, our best result of 85.47\% accuracy is highly competitive, placing it on par with or exceeding the zero-shot performance of much larger models.

It is particularly noteworthy that our fine-tuned 7B model surpasses the zero-shot performance of leading foundation models like the Llama-2-70B (85.0\%) \citep{touvron2023llama} and the original Llama-65B (85.3\%) \citep{touvron2023llama1}. This result strongly underscores the power of our synergistic fine-tuning approach. By efficiently composing weight-space and representation-space methods, we can unlock high-level reasoning capabilities in a relatively small model, achieving performance typically associated with models an order of magnitude larger. This highlights the potential of principled PEFT strategies to create highly capable, accessible, and resource-efficient specialized models.

\begin{table}[h!]
\centering
\caption{BoolQ validation accuracy across various models. Our fine-tuned 7B model outperforms many significantly larger base models evaluated in a zero-shot setting. Zero-shot performance numbers for base models are taken from \cite{touvron2023llama} and \cite{almazrouei2023falconseriesopenlanguage}.}
\label{tab:leaderboard_comparison}
\begin{tabular}{@{}llc@{}}
\toprule
\textbf{Model}  &  \textbf{Size}  &  \textbf{Accuracy (\%)}  \\  \midrule
\multicolumn{3}{c}{\textit{Zero-Shot Performance of Base Models}}  \\  \midrule
Falcon  &  7B  &  67.5  \\
&  40B  &  83.1  \\  \midrule
MPT  &  7B  &  75.0  \\
&  30B  &  79.0  \\  \midrule
Llama 1  &  7B  &  76.5  \\
&  13B  &  78.1  \\
&  33B  &  83.1  \\
&  65B  &  85.3  \\  \midrule
Llama 2  &  7B  &  77.4  \\
&  13B  &  81.7  \\
&  34B  &  83.7  \\
&  70B  &  85.0  \\  \midrule
PaLM  &  62B  &  84.8  \\  \midrule
Chinchilla  &  70B  &  83.7  \\
GPT-3  &  175B  &  60.5  \\
Gopher  &  280B  &  79.3  \\  \midrule
\multicolumn{3}{c}{\textit{Our Zero-Shot HEFT Results}}  \\  \midrule
HEFT (3 epochs)  &  7B  &  85.17  \\
HEFT (20 epochs)  &  7B  &  \textbf{85.47}  \\
\bottomrule
\end{tabular}
\end{table}

\section{Discussion}

The empirical success of our HEFT strategy provides strong support for the coarse-to-fine hypothesis, which can be understood through the lenses of curriculum learning, mechanistic synergy, and model modularity.

\paragraph{A methodological curriculum} Our approach can be framed as a form of implicit methodological curriculum learning. Unlike traditional curricula that sequence data from easy to hard, our approach sequences adaptation paradigms from broad to specific. The initial LoRA stage adapts the model to the general domain of the BoolQ task, moving the model into a parameter region well-suited for the required reasoning style. This creates a better-conditioned optimization landscape for the subsequent, more specialized ReFT stage, which can then converge more rapidly. This two-stage process mirrors patterns in human skill acquisition, where learning often involves first grasping general rules (a global adjustment, like LoRA) before refining specific, nuanced execution (a targeted adjustment, like ReFT).

\paragraph{A mechanistic perspective on synergy} The synergy can also be understood from a mechanistic perspective by considering the distinct operational domains of each method. LoRA, as a weight-space method, performs a global reparameterization. While effective, this process is indirect and has been shown to introduce structural artifacts, such as intruder dimensions, that can be causally linked to the forgetting of pre-trained knowledge \citep{sharma2024lora}. In contrast, ReFT originates from interpretability research showing that concepts are encoded within low-rank subspaces of the model's activation space \citep{wu2024reft}. We hypothesize that LoRA's coarse adaptation performs a broad alignment of the model's representation geometry, effectively moving these crucial semantic subspaces into a region more favorable for the task. The subsequent ReFT stage can then perform high-precision interventions directly within these subspaces, sharpening the exact computational pathways required for inference without the collateral effects of global weight updates. ReFT's surgical precision is thus amplified because it operates on a more amenable foundation prepared by LoRA.

\paragraph{Implications for model modularity} This hierarchical composition also introduces a novel perspective on model modularity. Much of the existing work on combining PEFT modules has focused on the parallel fusion of homogeneous adapters, such as averaging multiple LoRA modules \citep{asadi2024doescombiningparameterefficientmodules}. Our hierarchical approach suggests a different paradigm: creating specialized foundation models. The LoRA-adapted model is not just a temporary state but a new, consolidated base model with a foundational aptitude for a specific domain (in this case, inferential reasoning). On top of this new base, one can layer multiple, extremely lightweight ReFT interventions for even more specialized sub-skills, a concept that aligns with the goals of compositional methods like CS-ReFT \citep{pan2025csreft}.

\section{Limitations}
While our results are promising, this study has several limitations that provide context for our findings and suggest directions for future research.

\paragraph{Task and benchmark specificity} Our experiments are conducted exclusively on the BoolQ benchmark, a yes/no question-answering task focused on inferential reasoning. The strong synergistic effects we observe may not generalize equally to all other task types. For example, in long-form generative tasks, where weight-space methods like LoRA are often favored, the impact of a final ReFT stage might be different.

\paragraph{Order dependence} This work only investigates the hierarchical order of LoRA followed by ReFT, as motivated by our coarse-to-fine hypothesis. We did not explore the reverse order (ReFT-then-LoRA) or other potential compositions, such as an iterative, interleaved approach. The optimal ordering may be task-dependent and remains an open question.

\paragraph{Hyperparameter sensitivity} The configurations for LoRA (rank=8, alpha=32) and ReFT (layer=15, dimension=4) were chosen based on common practices in the literature. We did not perform an exhaustive hyperparameter search for either the individual methods or their hierarchical combination. It is possible that further tuning of these hyperparameters could yield different or even stronger results.

\section{Future work}
This research opens several avenues for future work. The immediate next step is to evaluate the HEFT strategy on a broader range of reasoning tasks (e.g., arithmetic, code generation, multi-hop QA) and across different model architectures and scales. Beyond direct replication, we propose three visionary directions:

\begin{enumerate}
  \item \textbf{Developing a PEFT algebra:} This work serves as a proof of concept for composing heterogeneous PEFT modules. Future research could explore a more general PEFT algebra, investigating the properties of different compositions (e.g., hierarchical, parallel, nested). This could lead to a principled framework for building complex model adaptations from a library of simple, efficient components.
  \item \textbf{Dynamic and conditional composition:} Instead of a fixed, static hierarchy, future methods could learn to dynamically compose PEFT modules. For instance, a lightweight routing network could decide whether to apply a ReFT intervention based on the input prompt or the model's internal state, leading to more efficient and context-aware models.
  \item \textbf{Generalizing the coarse-to-fine principle:} The principle of coarse-grained global adaptation followed by fine-grained local refinement could be a powerful recipe in other domains. For instance, in AI safety, one could use LoRA to instill broad safety principles and then use ReFT to surgically patch specific, nuanced vulnerabilities or jailbreaks, offering a more modular and interpretable approach to alignment.
\end{enumerate}

\section{Conclusion}
This paper introduced and validated HEFT (Hierarchical Efficient Fine-Tuning), a novel hierarchical fine-tuning strategy that combines weight-space adaptation (LoRA) with representation-space intervention (ReFT) in a coarse-to-fine manner. By first applying LoRA for broad adaptation followed by ReFT for precise refinement, we demonstrated a powerful synergistic effect on the challenging BoolQ inferential reasoning benchmark. Our primary finding is that this combined approach achieves superior accuracy with significantly less training time compared to using either method in isolation. Notably, a model trained for only three epochs with our hierarchical HEFT method outperformed models trained for a full 20 epochs with LoRA-only or ReFT-only.

This work provides compelling evidence that the principled composition of different PEFT methodologies is a highly promising direction for research and practice. It offers a practical and resource-efficient pathway to enhancing the reasoning capabilities of LLMs, directly contributing to the ongoing effort to build more efficient and powerful large reasoning models. The results suggest that the future of LLM adaptation may lie not in a single best method, but in the intelligent and hierarchical combination of complementary techniques.

\bibliographystyle{plainnat}
\bibliography{main}

\appendix
\section{Technical Appendices and Supplementary Material}
This appendix provides the Python script and HTCondor submission files used for the experiments.

\subsection{Python training and evaluation script}
\begin{lstlisting}[language=Python, caption={heft\_finetuning.py: Main script for HEFT (Hierarchical Efficient Fine-Tuning) and evaluation.}, label={lst:python_script}]
# Imports & helpers
import os, json, torch, transformers, pyreft
from peft import get_peft_model, LoraConfig, TaskType
from datasets import load_dataset
from huggingface_hub import login, HfApi
from torch.utils.data import Dataset
from tqdm import tqdm
try:
  from huggingface_hub import RepositoryNotFoundError
except ImportError:
  from huggingface_hub.utils import RepositoryNotFoundError

# Step 0: setup
os.environ["CUDA_VISIBLE_DEVICES"] = "0"
os.environ["PYTORCH_CUDA_ALLOC_CONF"] = "expandable_segments:True"
device = "cuda" if torch.cuda.is_available() else "cpu"
hf_token = os.getenv("HUGGING_FACE_HUB_TOKEN") or os.getenv("HF_TOKEN")
if not hf_token:
  raise ValueError("HUGGING_FACE_HUB_TOKEN env-var not set.")
login(token=hf_token)
BASE_MODEL   = "meta-llama/Llama-2-7b-chat-hf"
HF_USER      = "Bell-Herald" # Example user
# NOTE: Repo names would vary based on the experiment run (e.g., number of epochs)
HF_REPO_LORA = f"{HF_USER}/boolq_lora_example"
HF_REPO_REFT = f"{HF_USER}/boolq_reft_on_lora_example"
RESULTS_FILE = "evaluation_results.json"
api          = HfApi()

# Helper data & prompt
BOOLQ_PROMPT_TMPL = """<s>[INST] <<SYS>>
You are a helpful assistant that answers questions with only "Yes" or "No" based on the provided passage.
<</SYS>>
Passage: %s
Question: %s
Answer: [/INST]"""
def format_boolq(ex):
  prompt = BOOLQ_PROMPT_TMPL % (ex["passage"], ex["question"])
  answer = "Yes" if ex["answer"] else "No"
  return prompt, answer

# Stage 1: LoRA (Coarse-grained Adaptation)
print("\n Stage 1 - LoRA")
try:
  api.repo_info(HF_REPO_LORA)
  lora_exists = True
  print("Found LoRA repo on Hub.")
except RepositoryNotFoundError:
  lora_exists = False
  print("LoRA repo does not exist yet.")

if lora_exists:
  model = transformers.AutoModelForCausalLM.from_pretrained(
    HF_REPO_LORA, torch_dtype=torch.bfloat16, device_map=device)
  tokenizer = transformers.AutoTokenizer.from_pretrained(HF_REPO_LORA)
else:
  # Build LoRA from scratch
  model = transformers.AutoModelForCausalLM.from_pretrained(
    BASE_MODEL, torch_dtype=torch.bfloat16, device_map=device)
  tokenizer = transformers.AutoTokenizer.from_pretrained(
    BASE_MODEL, model_max_length=2048, padding_side="right", use_fast=False)
  tokenizer.pad_token = tokenizer.pad_token or tokenizer.eos_token

  boolq_train = load_dataset("boolq", split="train")
  dataset_lora = [format_boolq(ex) for ex in boolq_train]
  
  class SimpleDS(Dataset):
    def __init__(self, pairs, tok, max_len=2048):
      self.texts = [p + " " + a + tok.eos_token for p, a in pairs]
      self.tok = tok; self.max_len = max_len
    def __len__(self): return len(self.texts)
    def __getitem__(self, idx):
      enc = self.tok(self.texts[idx], truncation=True, max_length=self.max_len,
                     padding="max_length", return_tensors="pt")
      return {k: v.squeeze(0) for k, v in enc.items()}

  ds_lora  = SimpleDS(dataset_lora, tokenizer)
  collator = transformers.DataCollatorForLanguageModeling(tokenizer, mlm=False)
  lora_cfg = LoraConfig(task_type=TaskType.CAUSAL_LM, r=8, lora_alpha=32, lora_dropout=0.05)
  model   = get_peft_model(model, lora_cfg)
  
  # This value was varied for different experiments (e.g., 3 or 20)
  EPOCHS_LORA = 3 
  args_lora = transformers.TrainingArguments(
    output_dir="./lora_out", num_train_epochs=EPOCHS_LORA,
    per_device_train_batch_size=1, gradient_accumulation_steps=32,
    learning_rate=2e-4, evaluation_strategy="no", save_strategy="no",
    logging_steps=50, report_to="none"
  )
  trainer_lora = transformers.Trainer(
    model=model, tokenizer=tokenizer, args=args_lora,
    train_dataset=ds_lora, data_collator=collator
  )
  if EPOCHS_LORA > 0:
    trainer_lora.train()
  
  # Merge adapters and push to Hub
  model = model.merge_and_unload()
  api.create_repo(HF_REPO_LORA, exist_ok=True)
  model.push_to_hub(HF_REPO_LORA)
  tokenizer.push_to_hub(HF_REPO_LORA)
  print("LoRA artefact pushed.")

# Stage 2: ReFT (Fine-grained Refinement)
print("\nStage 2 - ReFT")
boolq_train = load_dataset("boolq", split="train")
pairs = [format_boolq(ex) for ex in boolq_train]
dm_reft = pyreft.make_last_position_supervised_data_module(
  tokenizer, model, [p for p, _ in pairs], [a for _, a in pairs])

reft_cfg = pyreft.ReftConfig(representations=[{
  "layer": 15, "component": "block_output", "low_rank_dimension": 4,
  "intervention": pyreft.LoreftIntervention(
    embed_dim=model.config.hidden_size, low_rank_dimension=4)
}])
reft_model = pyreft.get_reft_model(model, reft_cfg)

# This value was varied for different experiments (e.g., 3 or 20)
EPOCHS_REFT = 3 
args_reft = transformers.TrainingArguments(
  output_dir="./reft_out", num_train_epochs=EPOCHS_REFT,
  per_device_train_batch_size=8, gradient_accumulation_steps=4,
  learning_rate=2e-4, evaluation_strategy="no", save_strategy="no",
  logging_steps=50, report_to="none"
)
trainer_reft = pyreft.ReftTrainerForCausalLM(
  model=reft_model, tokenizer=tokenizer, args=args_reft, **dm_reft)

if EPOCHS_REFT > 0:
  trainer_reft.train()
print("ReFT artefact ready.")

# Step 3: Evaluation
print("\nStep 3 - Eval")
reft_model.set_device(device)
boolq_val = load_dataset("boolq", split="validation")
val_pairs = [format_boolq(ex) for ex in boolq_val]
correct = 0

print(f"Running on {len(val_pairs)} examples")
for prompt, truth in tqdm(val_pairs):
  enc = tokenizer(prompt, return_tensors="pt").to(device)
  base_unit_location = enc["input_ids"].shape[-1] - 1
  
  _, gen_ids = reft_model.generate(
    enc,
    unit_locations={"sources->base": (None, [[[base_unit_location]]])},
    intervene_on_prompt=True,
    max_new_tokens=5,
    eos_token_id=tokenizer.eos_token_id
  )
  pred_txt = tokenizer.decode(
    gen_ids[0, enc.input_ids.shape[1]:], skip_special_tokens=True).strip()
  pred = "Yes" if "Yes" in pred_txt else ("No" if "No" in pred_txt else "Unknown")
  correct += (pred == truth)
  
acc = 100 * correct / len(val_pairs)
print(f"Accuracy: {acc:.2f}%")

# Step 4: Save results
json.dump({
  "lora_epochs": EPOCHS_LORA,
  "reft_epochs": EPOCHS_REFT,
  "num_validation_samples": len(val_pairs),
  "correct_predictions": correct,
  "accuracy": f"{acc:.2f}%"
}, open(RESULTS_FILE, "w"), indent=4)
print(f"Saved results -> {RESULTS_FILE}")
\end{lstlisting}

\subsection{HTCondor submission files}
The following scripts were used to submit and run the jobs on the CHTC cluster.
\begin{lstlisting}[language=bash, caption={CHTC submit file (.sub)}, label={lst:sub_file}]
container_image = container.sif
executable = heft_exec.sh
transfer_input_files = heft_finetuning.py, .hf_token
should_transfer_files = YES
when_to_transfer_output = ON_EXIT
NAME = condor_outs/job_$(Cluster)_$(Process)
log = $(NAME).log
error = $(NAME).err
output = $(NAME).out
request_cpus = 1
request_gpus = 1
+WantGPULab = true
+GPUJobLength = "medium"
gpus_minimum_memory = 32000
Requirements = (Target.CUDADriverVersion >= 12.4)
request_memory = 32GB
request_disk = 32GB
queue 1
\end{lstlisting}

\begin{lstlisting}[language=bash, caption={Execution script (.sh)}, label={lst:sh_file}]
#!/bin/bash
echo "starting..."
HF_HOME="$(pwd)/.cache"
export HF_HOME
HF_TOKEN="$(cat .hf_token)"
export HF_TOKEN
WANDB_MODE="offline"
export WANDB_MODE

# Assuming a conda environment named 'reft' is available in the container
source activate reft

python heft_finetuning.py
echo "ending"
\end{lstlisting}

\end{document}